\documentclass[10pt,journal,compsoc]{IEEEtran}

\ifCLASSOPTIONcompsoc
  \usepackage[nocompress]{cite}
\else
  \usepackage{cite}  
\fi

\usepackage{subcaption}
\usepackage{graphicx}
\usepackage{amsmath}
\usepackage{amssymb}
\usepackage{booktabs}
\usepackage{multirow}
\usepackage{siunitx}
\usepackage{arydshln}

\usepackage{dblfloatfix}

\begin{document}
%
\title{TabSOM: A tabular-to-image encoding method based on self-organizing maps}

\author{David Chushig-Muzo,
        María Ángeles Rodríguez de Cara,
        Eva Milara,
        Francisco J. Lara-Abelenda,
        Luis Zhinin-Vera,
        Diego H. Peluffo-Ordóñez
\IEEEcompsocitemizethanks{\IEEEcompsocthanksitem David Chushig-Muzo, Eva Milara, María Ángeles Rodríguez de Cara are with the Department of Signal Theory and Communications, Telematics and Computing Systems, Rey Juan Carlos University (URJC), Madrid, Spain. E-mail: \{david.chushig, angeles.decara, eva.milara\}@urjc.es
\IEEEcompsocthanksitem Francisco J. Lara Abelenda is with the Faculty of Experimental Sciences, Universidad Francisco de Vitoria, Madrid, Spain. E-mail: francisco.lara@ufv.es
\IEEEcompsocthanksitem Luis Zhinin-Vera is with the LoUISE Research Group, University of Castilla-La Mancha, Albacete, Spain. E-mail: luis.zhinin@uclm.es
\IEEEcompsocthanksitem Diego H Peluffo-Ordoñez is with the School of Mathematical  and Computational Sciences, Yachay Tech University, Ecuador. E-mail: dpeluffo@yachaytech.edu.ec
}
}


\IEEEtitleabstractindextext{%
\begin{abstract}
Tabular-to-image methods have emerged as novel approaches to leverage the high predictive performance of convolutional neural networks and vision transformers. They convert tabular data into image representations, mapping each feature at a fixed pixel location derived from a dimensionality-reduction method (\textit{e.g.,} t-SNE, UMAP, PCA). However, they encode only the marginal value of each feature and discard information about feature relationships. We propose TabSOM, a tabular-to-image encoding built on the Self-Organizing Map (SOM), which provides: \textit{(i)} a spatial layout in which every input feature occupies a fixed canvas position derived from its component plane via collision-free Hungarian assignment; and \textit{(ii)} a graph that captures pairwise feature relationships derived from the SOM component planes. The resulting image stacks two multi-scale node channels: one encodes feature values at fixed scales, while the other encodes pairwise feature interactions as spatial connections between related features. Two SOM-derived interpretability approaches are introduced: a prototype-inspired partial dependence plot and a class-separation importance score. Benchmarked against twelve existing tabular-to-image methods across public binary-classification datasets, TabSOM ranks first or second on every dataset and achieves the lowest variance of any method evaluated. Interpretability obtained with TabSOM was validated against Random Forest, XGBoost, and SHAP, the class-separation score shows reasonable agreement with established baselines on the top-ranked features while capturing complementary structural information from input data. These results demonstrate that TabSOM provides an effective and interpretable approach for applying deep learning architectures to tabular data, bridging the performance–interpretability gap in this domain. 
\end{abstract}

}

\maketitle

\IEEEdisplaynontitleabstractindextext

\IEEEpeerreviewmaketitle

\ifCLASSOPTIONcompsoc
\IEEEraisesectionheading{\section{Introduction}\label{sec:introduction}}
\else
\section{Background and motivation}
\label{sec:introduction}
\fi

\IEEEPARstart{T}{abular} data are among the most widely used data formats for representing structured information~\cite{jiang2026representation}. It is characterized by a \textit{table-like format}, with rows corresponding to samples and columns to features. While Deep Learning (DL) models, such as Convolutional Neural Networks (CNNs) and Vision Transformers (ViTs), have achieved remarkable results on data characterized by high spatial or temporal correlations (\textit{e.g.,} images and audio)~\cite{mamdouh2026tab2visual}, tabular data generally exhibit weak interdependencies among features. As a result, DL architectures often show reduced effectiveness on predictive tasks involving tabular data.

This has fostered recent research into encoding methods, which convert tabular data into images to leverage the predictive performance of CNNs and pretrained ViTs~\cite{jiang2026representation}. Several tabular-to-image methods have been proposed in the literature, ranging from parametric to non-parametric approaches~\cite{liu2025comprehensive}. Most of these methods, such as DeepInsight~\cite{sharma2019deepinsight}, TINTO~\cite{castillo2023tinto}, REFINED~\cite{bazgir2020representation}, project features into two-dimensional through dimensionality reduction methods. They generate images in which spatial proximity reflects feature similarity, with neighboring pixels representing features with some similarity. Tabular-to-image methods have demonstrated competitive performance in several domains such as healthcare~\cite{lara2025transfer, singh2026enhanced, gomez2026tabular}, indoor localization~\cite{liu2026interpretable, castillo2025mimo, talla2023novel}, and the internet of things~\cite{babayigit2026image} among others.

The performance of tabular-to-image encoding methods directly depends on how features are spatially arranged and how their values are encoded. Existing methods have addressed this mainly through dimensionality reduction methods~\cite{castillo2023tinto, sharma2019deepinsight}, including Principal Component Analysis (PCA), kernel PCA, t-distributed Stochastic Neighbor Embedding (t-SNE) and Uniform Manifold Approximation and Projection (UMAP). These project the feature space into two-dimensional space, assigning each feature to its nearest pixel on the image canvas. Although they produce effective spatial layouts, the embedding is used solely to determine feature locations. As a result, each feature occupies a fixed position derived from the feature distribution, without explicitly encoding relationships among feature values.

The Self-Organizing Map (SOM), an artificial neural network based on unsupervised and competitive learning, performs a topology-preserving mapping from the input space to a two-dimensional grid of nodes~\cite{kohonen1996engineering}. The topology-preserving property enables that similar inputs are mapped to neighbor nodes on the grid. Then, local relationships in the input space are reflected in spatial proximity on the grid. As a result, SOM has been extensively used for visualization, clustering, and pattern recognition in different domains such as signal processing, healthcare among others~\cite{chushig2020data, chon2011self, astudillo2014topology}. Although SOM has shown excellent results in unsupervised tasks, its application within tabular-to-image methods has been partially explored~\cite{achutha2025topological}. Unlike other dimensionality reduction techniques, the SOM preserves an interpretable structure. Each node is associated with a prototype, the grid of prototypes provides a density-weighted summary of the joint feature distribution, and the component planes capture how each feature varies across the map. Although tabular-to-image encoding methods have shown promising predictive performance, their interpretability has received limited attention. The graph structure of the SOM captures the topology of the data while providing interpretability.

In this paper, we propose a tabular-to-image encoding method named TabSOM, which includes pairwise feature interactions as part of the image. We train a SOM on the feature space and extract, for each feature, its component plane, \textit{i.e.,} the spatial distribution of that feature's values across the trained SOM grid. We derive a canvas position (named anchor) for each feature from its component plane, and then optimize anchor collisions via a Hungarian assignment to grid cells. The SOM's prototype grid serves as a density-weighted, topologically organized sample of the joint feature distribution, and the CNN's predictions on those prototypes define a spatial prediction map over the grid. One of the main advantages of TabSOM over other tabular-to-image methods is that the SOM is not used solely to fit the layout but also provides two interpretability methods derived from the grid: \textit{(i)} global feature ranking; and \textit{(ii)} prototype-inspired partial dependence curves. 
 
The main contributions of this paper are the following:

\begin{itemize}
    \item A novel tabular-to-image encoding method in which feature placement is derived from SOM component planes via anchor estimation and collision-free (Hungarian) assignment.
    
    \item A relational channel that renders pairwise feature interactions derived from the SOM prototype geometry that determines placement. 
    
    \item A multi-scale rendering scheme in which the same feature layout is rendered at multiple Gaussian bandwidths as aligned channels, providing a CNN with both precise localization and regional coherence without requiring a single bandwidth choice.

    \item An SOM-based interpretability framework that leverages the learned grid structure to provide two complementary explanations: \textit{(i)} global feature ranking; and \textit{(ii)} prototype-inspired partial dependence curves.

    \item Feature maps that allow the encoding's class-separability and per-feature spatial assignment to be inspected independently of any downstream classifier.
        
\end{itemize}

The remainder of this paper is organized as follows. Section~\ref{sec:related_work} reviews tabular-to-image encoding methods of the literature. Section~\ref{sec:methods} presents each component of TabSOM and its derived interpretability tools. Section~\ref{sec:results} describes the datasets used, experimental setup, and main results, including classification benchmark and interpretability analysis. Section~\ref{sec:discussion} discusses limitations and directions for future work, and Section~\ref{sec:conclusions} presents the main conclusions.

\section{Related work}
\label{sec:related_work}

A range of tabular-to-image methods have been proposed in the literature, which are categorized into two main categories~\cite{liu2025comprehensive}: non-parametric and parametric methods. 

Non-parametric methods use heuristic algorithms to map features into a predefined image-grid representation without explicitly optimizing the spatial arrangement of features~\cite{liu2025comprehensive}. Representative approaches include the following. BarGraph encodes each tabular sample as a vertical bar chart, where each feature is assigned to a fixed column and the bar height corresponds to the normalized feature value~\cite{sharma2022classification}. Binary Image Encoding (BIE) and correlated BIE convert each numerical value to binary strings, then stack these bit-rows to form a two-dimensional matrix where 0 and 1 denote dark and light pixels, respectively~\cite{briner2023tabular}. SuperTML encodes feature values as text on an image, using CNNs as character/word-level feature extractors~\cite{sun2019supertml}. Tab2Visual encodes each feature as a vertical bar whose width is proportional to the normalized feature value, arranged in rows and columns of bars~\cite{mamdouh2025tab2visual}. 

Parametric methods perform the spatial arrangement of features through an optimization-based preprocessing stage. Linear and nonlinear dimensionality reduction techniques (such as PCA, t-SNE, and UMAP) project features from the input space into a low-dimensional coordinate space, with the resulting coordinates defining the feature layout from which the image is built. Among the most popular methods, we find DeepInsight~\cite{sharma2019deepinsight} and REFINED~\cite{bazgir2020representation}, which use t-SNE, kernel PCA and Multidimensional Scaling to obtain a 2D position for each feature, and then map feature values to pixel intensities. TINTO uses PCA and t-SNE to determine feature locations  and the resulting coordinates are then transposed, scaled, and rounded to integer values~\cite{castillo2023tinto}. It incorporates a blurring mechanism to produce smoother image representations, thereby enhancing compatibility with convolutional filters~\cite{castillo2023tinto}. 

Fotomics applies the Fourier Transform independently to each feature column and represents the resulting complex coefficients through their real and imaginary components as coordinates in a two-dimensional Cartesian plane~\cite{alenizy2025transforming}. FC‑Viz aggregates highly correlated features into clusters and evaluates the relationships between representative features from each cluster~\cite{damri2024towards}. Ant colony optimization determines the spatial organization of the features, while dimensionality reduction methods are employed to compute pixel intensities~\cite{damri2024towards}. Other methods in this category are IGTD~\cite{zhu2021converting} and LM-IGTD~\cite{gomez2024lm}. They perform a feature-to-pixel assignment as a distance-preservation optimization problem, minimizing the difference feature similarities and distances between samples. LM-IGTD~\cite{gomez2024lm} extends IGTD to handle low-dimensional and mixed-type data through an unsupervised stochastic noise generation. 

Table~\ref{table:summary_tabular_image_methods} presents a summary of tabular-to-image methods, including category and technique used for the image layout.

\begin{table}[!ht]
\centering
\caption{Summary of tabular-to-image encoding methods, including their category and the technique used to construct the image layout. Methods are shown in alphabetical order.}
\label{table:summary_tabular_image_methods}
\scalebox{0.8}{
\begin{tabular}{@{}llc@{}}
\toprule
Method & Category & Technique for image layout \\
\midrule
DeepInsight~\cite{sharma2019deepinsight}   & Parametric  & t-SNE \\
Fotomics~\cite{alenizy2025transforming}    & Parametric  & Fourier Transform \\
FC-Viz~\cite{damri2024towards}             & Parametric  & t-SNE, KPCA, UMAP and colony optimization \\
IGTD~\cite{zhu2021converting}              & Parametric  & distance-preservation optimization \\
LM-IGTD~\cite{gomez2024lm}                 & Parametric  & distance-preservation optimization \\
REFINED~\cite{bazgir2020representation}    & Parametric  & Bayesian Multidimensional Scaling  \\
TINTO~\cite{castillo2023tinto}             & Parametric  & PCA, t-SNE \\  \cdashline{1-3}
BarGraph~\cite{sharma2022classification}   & Non-parametric & -- \\
BIE~\cite{halladay2025bie}                 & Non-parametric & -- \\
SuperTML~\cite{sun2019supertml}            & Non-parametric & -- \\
Tab2Visual~\cite{mamdouh2025tab2visual}    & Non-parametric & -- \\
\bottomrule
\end{tabular}
}
\end{table}

\section{Methods}
\label{sec:methods}

\subsection{TabSOM}

TabSOM consists of five stages: (1) SOM training and component-plane extraction, (2) feature anchor estimation, (3) collision-free spatial placement, (4) feature relationship graph construction, and (5) multi-channel image rendering. Each stage is described in detail below. 

\subsubsection{Self-organizing map training}

Let $X \in \mathbb{R}^{N \times F}$ be a tabular dataset with $N$ samples and $F$ features, and let $y \in \{0,1\}^{N}$ denote the corresponding vector labels. Each feature is first scaled using min-max or $z$-score normalization. A rectangular SOM with $H \times W$ nodes is trained on the normalized data using the standard online SOM update rule. Each node $(r,c)$ has associated a prototype vector $\mathbf{w}_{r,c} \in \mathbb{R}^F$, initialized by sampling samples from the training data. At each training step $t$, a sample $\mathbf{x}$ is drawn at random, and its best matching unit (BMU) is identified as

$$
b = \arg\min_{(r,c)} \lVert \mathbf{w}_{r,c} - \mathbf{x} \rVert^2,
$$

and every node's prototype is updated according to the rule:

$$
\mathbf{w}_{r,c} \leftarrow \mathbf{w}_{r,c} + \eta(t) \, h_{\sigma(t)}(r,c,b) \,
\left( \mathbf{x} - \mathbf{w}_{r,c} \right),
$$

where the learning rate $\eta(t)$ and neighborhood radius $\sigma(t)$ decay exponentially from initial values $\eta_0, \sigma_0$ to final values $\eta_1, \sigma_1$ over the course of training, and $h_\sigma(r,c,b) = \exp\left(
-\frac{(r-r_b)^2+(c-c_b)^2}{2\sigma^2} \right)$ is a Gaussian neighborhood function over grid coordinates. The grid size $H \times W$ is chosen such that the number of cells $K = H \cdot W$ is at least $F$, ensuring sufficient placement capacity for features.

\subsubsection{Component planes and feature anchors}

After training, the $j$-th component plane $\Phi_j \in \mathbb{R}^{H \times W}$ is defined as the $j$-th coordinate of every node's prototype vector, $\Phi_j(r,c) = \left[ \mathbf{w}_{r,c} \right]_j $, a smooth spatial field over the grid whose value at $(r,c)$ indicates how high feature $j$ tends to be among samples mapped near that node. 

Because the SOM update rule pulls neighboring nodes toward similar prototypes, $\Phi_j$ varies gradually across the grid rather than at random. Nodes that are close together on the map represent similar combinations of feature values, then $\Phi_j$ exhibits a coherent region of high intensity rather than several disconnected peaks. This topology preservation is what allows a single scalar feature to be associated with a location on the grid in the first place, rather than only with a value at
each node independently.

We exploit this property to derive, for every feature, a canvas position that reflects where that feature is characteristically expressed across the training distribution. Each feature $j$ is first assigned an anchor position ($\mathbf{a}_j \in [0,1]^2$), representing its preferred location on the image canvas before collision resolution. 

Component planes form the basis for both feature placement and the relational graph. Each feature $j$ is assigned an anchor position $\mathbf{a}_j \in [0,1]^2$, its preferred location on the image canvas, computed from $\Phi_j$ by one of two methods:

\textit{Centroid anchor.} The intensity-weighted center of mass of the (optionally sharpened) component plane:
$$
\mathbf{a}_j = \frac{\sum_{(r,c)} \mathbf{g}(r,c) \cdot \tilde\Phi_j(r,c)^{\,\gamma}}
{\sum_{(r,c)} \tilde\Phi_j(r,c)^{\,\gamma}},
$$

where $\tilde\Phi_j$ is $\Phi_j$ normalized to $[0,1]$, $\mathbf{g}(r,c) \in [0,1]^2$ is the normalized grid coordinate of cell $(r,c)$, and $\gamma \geq 1$ is a sharpening exponent (default $\gamma=2$) that pulls the anchor toward the plane's peak by down-weighting low-intensity regions before averaging.

\textit{Mode anchor.} The location of the plane's maximum, $\arg\max_{(r,c)} \Phi_j(r,c)$, is optionally refined to sub-pixel precision via parabolic interpolation over the peak's immediate neighborhood along each axis. Mode anchors better separate features whose component planes have similar overall mass but different peak locations, at the cost of anchors clustering toward canvas edges/corners for sharply peaked planes.

In practice, the centroid anchor is preferred when component planes are broad and overlapping (typical of low-to-moderate $F$, where the SOM has ample grid capacity per feature), while the mode anchor is preferred when $F$ is large and many planes compete for similar grid regions, since a small shift in peak location is then more informative for separating features than a shift in their overall mass. Both anchor methods operate purely on $\Phi_j$ and therefore on a fixed, $N$-independent $H \times W$ grid, so the cost of computing all $F$ anchors does not grow with the number of samples once the SOM itself has been trained.

\subsubsection{Collision-free feature placement}

Anchors computed may coincide or be placed closely, which would cause multiple features to overlap on the rendered image canvas. To solve this, we apply a discrete optimal-assignment step. Let $\{g_1, \ldots, g_K\}$ denote the centers
of the $K = H \times W$ canvas cells (in normalized $[0,1]^2$ coordinates). The cost of placing feature $j$ in cell $c$ is

$$
C_{jc} = \alpha \, \lVert g_c - \mathbf{a}_j \rVert - \beta \, \tilde\Phi_j(g_c),
$$

where the first term penalizes displacement from the feature's anchor and the second (optional, $\beta \geq 0$) rewards placing the feature where its own component plane is locally strong. The final placement $\sigma: \{1,\ldots,F\} \to \{1,\ldots,K\}$ is obtained by solving the linear assignment problem

$$
\sigma^* = \arg\min_{\sigma} \sum_{j=1}^{F} C_{j,\sigma(j)}
$$

via the Hungarian algorithm, guaranteeing that no two features share a cell (zero overlap) while minimizing total displacement from the SOM-derived anchors. Feature $j$'s final canvas position is the center of cell
$\sigma^*(j)$.

\subsubsection{Feature relationship graph}

To capture pairwise feature interactions, we construct a feature relationship graph directly from the component planes. Each component plane $\Phi_j$ is flattened to a vector in $\mathbb{R}^{HW}$, and the relationship weight between features $i$ and $j$ is computed as either their Pearson correlation or cosine similarity across the flattened planes:

$$
W_{ij} = \text{corr}\!\left(\text{vec}(\Phi_i), \text{vec}(\Phi_j)\right)
$$

$$
W_{ij} = \frac{\text{vec}(\Phi_i)^\top \text{vec}(\Phi_j)}
{\lVert \text{vec}(\Phi_i) \rVert \, \lVert \text{vec}(\Phi_j) \rVert}.
$$

Negative weights are clipped to zero (retaining only positive co-variation), and the diagonal is set to zero. The edge set $E$ is obtained by thresholding: $E = \{ (i,j) : W_{ij} \geq \tau \}$ for a threshold $\tau$
(default $0.5$–$0.6$). Because $W$ is derived from the same prototype geometry that determines feature placement, features connected by an edge are typically placed near each other on the canvas.

\subsubsection{Multi-channel image rendering}

Given the placement $\{\mathbf{p}_j\}_{j=1}^F \subset [0,1]^2$ (canvas positions) and the edge set $E$ with weights $\{W_{ij}\}_{(i,j) \in E}$, a tabular sample $\mathbf{x} \in \mathbb{R}^F$ is rendered into an image $I \in \mathbb{R}^{H_{\text{img}} \times W_{\text{img}} \times C}$ with $C$ channels. We selected $C=3$, with two first node channels and a relational edge channel.

\textit{Node channels}. For each of $S$ Gaussian bandwidths $\sigma_1 < \sigma_2 < \cdots < \sigma_S$ (default $S=3$, e.g. $\sigma \in \{0.05, 0.10, 0.18\}$, sharp to smooth), one channel is rendered as

$$
I_s(u,v) = \sum_{j=1}^{F} x_j \cdot \exp\!\left( -\frac{\lVert (u,v) -
\mathbf{p}_j \rVert^2}{2\sigma_s^2} \right), \qquad s = 1,\ldots,S,
$$

where $(u,v)$ ranges over pixel centers in $[0,1]^2$. Each channel is the same spatial layout rendered at a different bandwidth: the sharp channel localizes each feature precisely (supporting fine-grained discrimination), while the smooth channel provides regional coherence over a neighborhood the size of a convolutional kernel's receptive field (supporting the local-pattern assumptions of CNNs). Because all $S$ channels share the same feature positions, they are spatially aligned, allowing a CNN to jointly exploit multiple scales — analogous to a feature pyramid. We use $S=2$ ($\sigma \in \{0.05, 0.08\}$), reserving the third channel for the relational edge channel.

\textit{Relational edge channel}. For each edge $(i,j) \in E$, define the segment field as the Gaussian distance from $(u,v)$ to the line segment connecting $\mathbf{p}_i$ and
$\mathbf{p}_j$:

$$
\text{seg}_{ij}(u,v) = \exp\!\left( -\frac{ d\big((u,v),\ \overline{\mathbf{p}_i
\mathbf{p}_j}\big)^2 }{2\sigma_e^2} \right),
$$

where $d(\cdot,\cdot)$ is the perpendicular distance to the segment (clamped to the segment's endpoints) and $\sigma_e$ is a fixed edge bandwidth. The edge channel is then

$$
I_{\text{edge}}(u,v) = \sum_{(i,j) \in E} W_{ij} \cdot \sqrt{\max(x_i,0)\cdot
\max(x_j,0)} \cdot g_{ij}(\mathbf{x}) \cdot \text{seg}_{ij}(u,v),
$$

where $\sqrt{x_i x_j}$ is the interaction intensity — non-zero only when both endpoint features are active for this sample — scaled by the relationship weight $W_{ij}$. This channel is therefore a spatial map of which pairwise feature interactions are active for the specific row being encoded, distinct from the node channels, which encode only marginal feature values.

The final image stacks all channels, $I = [I_1, \ldots, I_S, I_{\text{edge}}\,
(, I_{\text{graph}})]$, with $C = S + 1$ (or $S+2$ with the graph channel); in our primary configuration ($S=2$, edge channel included, no static graph channel) this gives $C=3$. 

\subsection{TabSOM-derived interpretability analysis}

To provide interpretability, we derive two methods from the trained SOM, which provide a two-dimensional representation of the joint feature distribution.

Let the trained SOM consist of a grid of $K = H \times W$ nodes. Each node $(r,c)$ has a prototype weight vector $w_{r,c} \in [0,1]^F$, where $F$ is the number of input features. The component plane of feature $j$ is $\Phi_j(r,c) = w_{r,c,j}$, i.e. the $j$-th coordinate of the prototype at node $(r,c)$. For each training sample $x_i$, its best-matching unit $\mathrm{BMU}(x_i) \in \{1,\dots,H\} \times \{1,\dots,W\}$ is the node whose prototype is closest in Euclidean distance.

\subsubsection{Self-organizing map class-separation importance}

For binary classification, we define class-conditional activation densities over the SOM grid from the BMU assignments of the training set. For class $k \in \{0,1\}$,

$$
A_k(r,c) = \frac{\big|\{i : y_i = k,\ \mathrm{BMU}(x_i) = (r,c)\}\big|}{\big|\{i : y_i = k\}\big|},
$$

$A_1$ and $A_0$ are normalized satisfying $\sum_{r,c} A_k(r,c) = 1$. The class-separation importance of feature $j$ is the absolute difference between the $A_1$- and $A_0$-weighted averages of its component plane,

$$
\mathrm{Imp}^{\text{class-sep}}_j = \left| \sum_{r,c} A_1(r,c)\,\Phi_j(r,c) - \sum_{r,c} A_0(r,c)\,\Phi_j(r,c) \right|.
$$

This quantity is large when feature $j$ takes systematically different typical values in the map regions favored by each class. It is computed from the SOM together with the (binary) training labels, independent of  downstream predictive models. We use $\mathrm{Imp}^{\text{class-sep}}_j$ as the primary SOM-derived global feature-importance ranking.

\subsubsection{Prototype-based partial dependence plot}

Each SOM prototype $w_{r,c}$ represents a density-weighted combination of feature values drawn from the trained SOM's organization of the input space. In contrast to the synthetic, independence-assuming Cartesian grids used by conventional Partial Dependence Plots (PDPs). We exploit this by encoding every prototype with the fitted encoder and passing the resulting image through the trained CNN $f_\theta$, yielding a prediction map

$$
\Psi(r,c) = f_\theta\big(\mathrm{Encode}(w_{r,c})\big) \in [0,1],
$$

interpreted as the predicted probability $P(y=1)$ for the (synthetic but data-consistent) sample represented by node $(r,c)$. $\Psi$ has the same spatial shape $(H,W)$ as every component plane $\Phi_j$.

For a given feature $j$, we construct a prototype-based PDP by plotting the pairs $\big(\Phi_j(r,c), \Psi(r,c)\big)$ over all $K$ nodes, where $\Phi_j(r,c)$ is mapped back to the feature's original units via the inverse min-max transform. Each point is weighted by its hit count

$$
n(r,c) = \big|\{i : \mathrm{BMU}(x_i) = (r,c)\}\big|,
$$

the number of training samples for which node $(r,c)$ is the best-matching unit. A density-weighted mean curve is obtained by partitioning the nodes into $B$ bins of approximately equal cumulative hit count, ordered by $\Phi_j$, and computing the hit-count-weighted mean of $\Psi$ within each bin. This curve is read analogously to a standard PDP — the expected model output as a function of feature $j$ — but is restricted to regions of feature space the SOM actually populates with prototypes, so that sparsely-supported (low hit-count) segments are visually distinguishable from well-supported ones. Where a prototype-based PDP is non-monotonic, the underlying spatial maps $\Phi_j$ and $\Psi$ can be inspected directly on the shared $(H,W)$ grid to identify whether the non-monotonicity arises from an interaction with another feature (\textit{i.e.,} a region of the grid where some other component plane, rather than $\Phi_j$, aligns with $\Psi$).

\section{Results}
\label{sec:results}

\subsection{Datasets}
\label{sec:dataset_description}

We evaluate TabSOM on four real-world datasets: Oxford Parkinson's Disease (PAR), Pima Indians Diabetes (PID), QSAR Biodegradation (QSA), and Wisconsin Diagnostic Breast Cancer (WBC). All datasets are obtained from the public repository UCI Machine Learning Repository~\footnote{UCI Repository: http://archive.ics.uci.edu/ml/} and present two classes (binary classification). Table~\ref{table:dataset_summary} summarizes the datasets, including the total number of samples and features. 

\begin{table}[htbp]
\centering
\caption{Datasets considered in this study.}
\label{table:dataset_summary}
\scalebox{0.95}{
\begin{tabular}{llrr}
\toprule
Identifier   & Dataset                          & Samples  & Features  \\
\midrule
PID  & Pima Indians Diabetes            & 768  & 8                \\
PAR  & Oxford Parkinson's Disease       & 195  & 22                \\
QSA  & QSAR Biodegradation              & 1055 & 41                \\
WBC  & Wisconsin Breast Cancer          & 569  & 30                \\
\bottomrule
\end{tabular}
}
\end{table}

\subsection{Experimental setup}
\label{sec:experimental_setup}

We split each dataset into two independent subsets, a training subset (80\% samples) and test subset (20\% samples). To evaluate the generalization of predictive models, all methods are evaluated under 5-fold stratified cross-validation. Class imbalance was addressed through random undersampling, and it is applied only for training subset to prevent data leakage. All features are min-max normalized to $[0,1]$. We evaluate the predictive performance using the Area Under the Receiver Operating Characteristic Curve (AUROC). Results are reported as the mean and standard deviation across five random seeds.

TabSOM is configured as follows. The SOM grid side is set automatically to $H = W = \max(8, \lceil\sqrt{1.3 F}\rceil)$. Feature placement uses Hungarian assignment with centroid anchors, and the relational graph uses Pearson correlation between component planes with threshold $\tau = 0.5$. The primary rendering configuration uses $S = 2$ node channels at $\sigma \in \{0.05, 0.08\}$ and the relational edge channel.

Encoded images are classified using a CNN with three convolutional blocks (16-32-64 channels), each followed by batch normalization and ReLU, with a max-pooling step after the second block — followed by global average pooling and a single linear output unit). The network is trained with Adam, the learning rate $10^{-3}$, weight decay $10^{-4}$, batch size of 32 and a class-balanced binary cross-entropy loss. The same architecture and optimizer are used for every tabular-to-image encoding method.

\subsection{Classification results and benchmark}

\begin{table*}[!t]
\centering
\caption{Benchmark of tabular-to-image methods using mean and standard deviation AUCROC values across five random seeds.}
\label{table:benchmark_t2i_methods}
\begin{tabular}{lcccc}
\toprule
Method & Pima & Parkinsons & WDBC & QSAR \\ \midrule

BarGraph &
$0.8100 \pm 0.0301$ &
$0.8753 \pm 0.0297$ &
$0.9859 \pm 0.0055$ &
$0.8876 \pm 0.0162$ \\

BIE &
$0.7585 \pm 0.0513$ &
$0.7321 \pm 0.0810$ &
$0.9765 \pm 0.0140$ &
$0.8455 \pm 0.0259$ \\

Combination &
$0.8199 \pm 0.0285$ &
$\mathbf{0.9179 \pm 0.0312}$ &
$0.9897 \pm 0.0046$ &
$0.9179 \pm 0.0163$ \\

DeepInsight (tSNE) &
$0.5898 \pm 0.0835$ &
$0.7134 \pm 0.2593$ &
$0.7576 \pm 0.1433$ &
$0.5636 \pm 0.1207$ \\

DeepInsight (UMAP) &
$0.5925 \pm 0.1067$ &
$0.6234 \pm 0.1948$ &
$0.6927 \pm 0.2327$ &
$0.8231 \pm 0.0526$ \\

DistanceMatrix &
$0.7150 \pm 0.0532$ &
$0.8993 \pm 0.0480$ &
$0.9624 \pm 0.0170$ &
$\mathbf{0.9140 \pm 0.0136}$ \\

FeatureWrap &
$0.7839 \pm 0.0425$ &
$0.7776 \pm 0.1256$ &
$0.9618 \pm 0.0135$ &
$0.8392 \pm 0.0433$ \\

Fotomics &
$0.5615 \pm 0.0994$ &
$0.6842 \pm 0.2643$ &
$0.6349 \pm 0.3296$ &
$0.7612 \pm 0.0829$ \\

IGTD &
$0.5548 \pm 0.1190$ &
$0.8565 \pm 0.0321$ &
$0.7826 \pm 0.1759$ &
$0.5887 \pm 0.0701$ \\

TINTO (PCA) &
$0.6001 \pm 0.0755$ &
$0.8472 \pm 0.0815$ &
$0.7881 \pm 0.0971$ &
$0.7955 \pm 0.0438$ \\

TINTO (tSNE) &
$0.5199 \pm 0.0957$ &
$0.7761 \pm 0.1251$ &
$0.6770 \pm 0.1353$ &
$0.7076 \pm 0.1109$ \\

TINTO (tSNE and blur) &
$0.5265 \pm 0.0694$ &
$0.6730 \pm 0.2401$ &
$0.7341 \pm 0.1120$ &
$0.6540 \pm 0.0957$ \\

\textbf{TabSOM} &
$\mathbf{0.8236 \pm 0.0240}$ &
$0.8852 \pm 0.0390$ &
$\mathbf{0.9911 \pm 0.0082}$ &
$0.9098 \pm 0.0180$ \\

\bottomrule
\end{tabular}
\end{table*}

Table~\ref{table:benchmark_t2i_methods} compares TabSOM against twelve existing tabular-to-image encoding methods across six binary datasets. Across the four datasets, TabSOM ranks first in AUCROC on Pima (0.8236) and WDBC (0.9911), and third on PAR (0.8852) and QSAR (0.9098). These results yield the second-highest overall mean AUCROC (0.9024) and the second-best average rank, with only the Combination method performing better overall (mean AUCROC: 0.9114). The performance gap between TabSOM and the strongest competing method is small across all datasets. TabSOM outperforms the next-best method (Combination) by 0.0037 AUCROC on Pima and 0.0014 on WDBC, while underperforming Combination by 0.0327 AUCROC on Parkinsons and 0.0081 on QSAR. This places TabSOM and Combination as the two clearly strongest methods in the benchmark, consistently separated from the rest of the field by a substantial margin — the third-place method on most datasets (BarGraph or DistanceMatrix) trails TabSOM by roughly 0.01–0.04 AUCROC, while methods based on generic dimensionality-reduction embeddings (TINTO (tSNE), DeepInsight (tSNE), DeepInsight (UMAP), Fotomics) trail by considerably more, often falling 0.15–0.3 AUC below TabSOM and showing markedly higher variance (standard deviations frequently exceeding 0.10, compared to TabSOM's 0.018–0.039 across all four datasets). 

Additionally, TabSOM's standard deviation is among the lowest of any method on every dataset, indicating more stable performance across different seeds than most tabular-to-image approaches. Several methods (TINTO (tSNE and blur) on PAR, DeepInsight (tSNE) on PAR, Fotomics on WDBC) show standard deviations an order of magnitude larger, suggesting these embeddings are less reliable across different seeds. Finally, the gap between TabSOM and the weaker methods widens on datasets with fewer features (Pima, 8 features), where structured encodings (TabSOM, BarGraph, Combination, DistanceMatrix) outperform embedding-based methods by the largest margin (\textit{e.g.,} TabSOM exceeds TINTO (tSNE) by 0.30 AUCROC on Pima), while on higher-dimensional or more separable datasets (WDBC) most methods cluster more closely near the best performance.

\subsection{Channel decomposition and feature maps}

\begin{figure*}[h]
    \centering
    \includegraphics[width=0.95\textwidth]{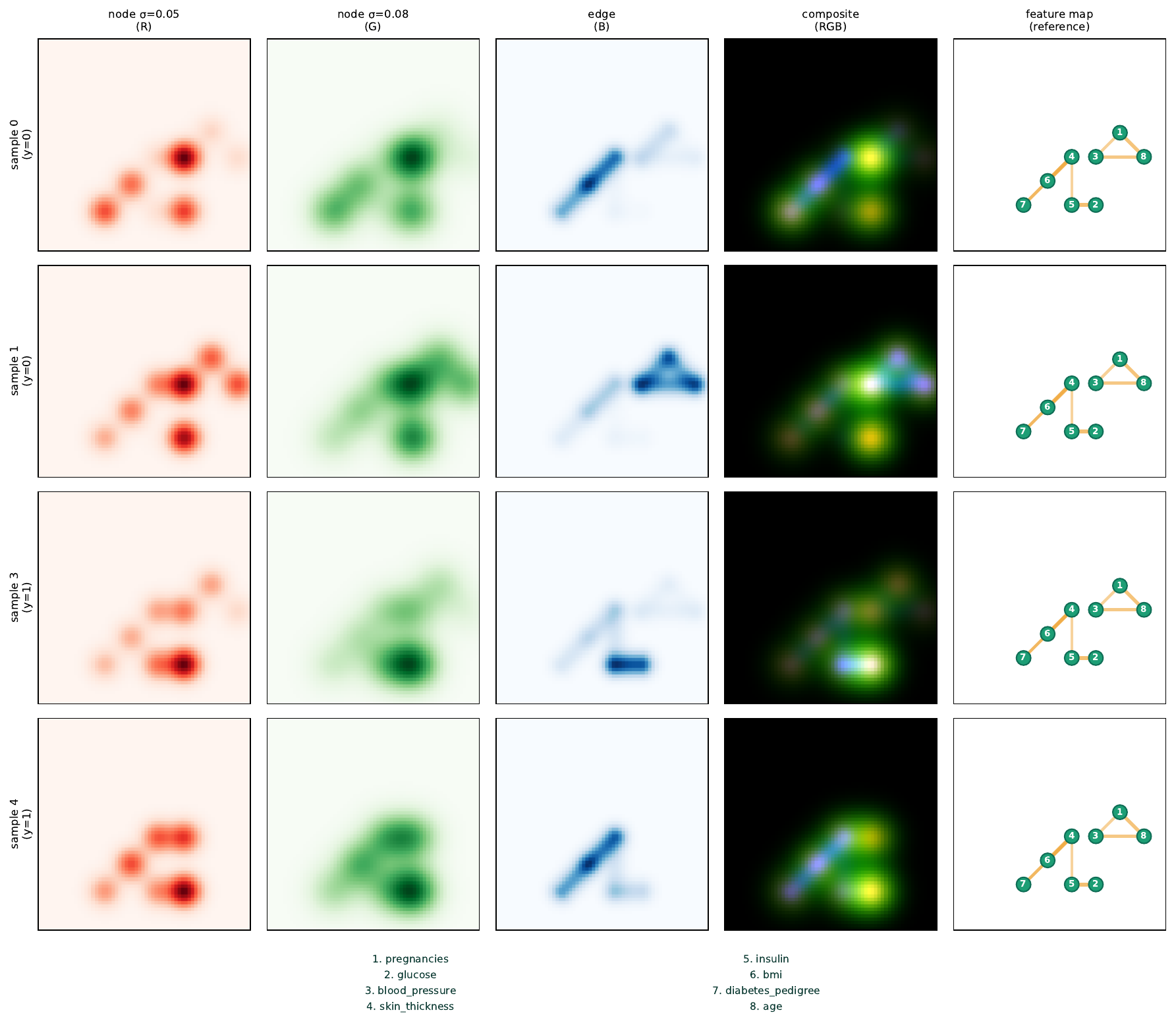} 
    \caption{Channel decomposition of the TabSOM encoding for four Pima Indians Diabetes samples. Three first columns show the individual channels: the sharp node channel (R, $\sigma$=0.05) and mid node channel (G, $\sigma$=0.08) render each feature's value as a Gaussian field centered at its SOM-assigned position, while the edge channel (B) renders the relational graph's pairwise feature interactions. Fourth channel shows the composite RGB image, whereas the fifth column presents the feature map.}
    \label{fig:feature_maps}
\end{figure*}

Figure~\ref{fig:feature_maps} shows channel decomposition and final image resulting of the TabSOM encoding for samples from the PID. The sharp and mid node channels preserve each feature's individual contribution at a stable canvas position across samples, with intensity tracking the feature's normalized value. The edge channel shows the pairwise relationships captured by the correlation-based relational graph. A comparison across samples indicates that the overall structure of these channels is consistent among samples, whereas localized intensity differences indicate patient-specific feature values. Regarding feature relationships, for instance, sample 3 exhibits substantially stronger activation along the blood\_pressure–age and skin\_thickness–blood\_pressure edges than the other samples. This pattern is evident both in the edge channel and in the composite image, where it appears as a pronounced cyan–white region.

\subsection{SOM-derived feature importance and dependence analysis}

\begin{figure*}
  \centering
  \begin{subfigure}[b]{0.9\linewidth}
    \includegraphics[width=\linewidth]{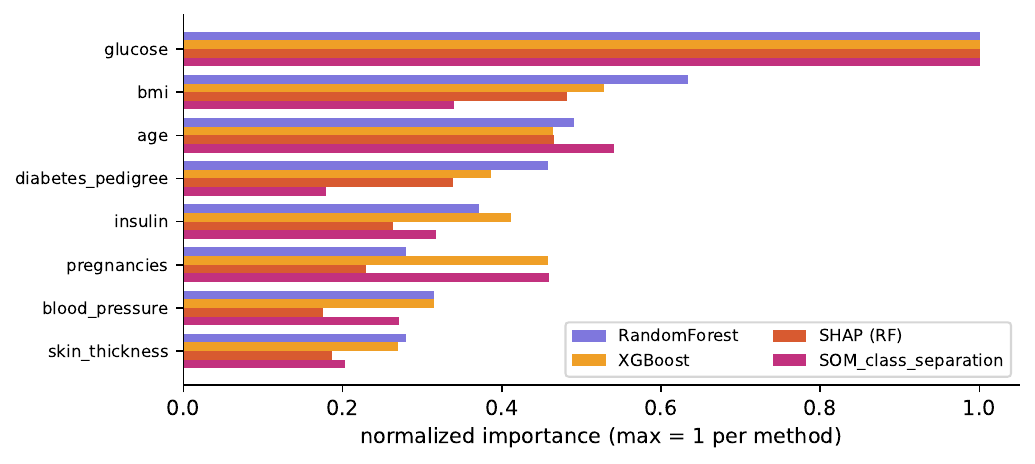} 
    \caption{} 
    \label{fig:gradcam_local:a} 
  \end{subfigure}
  \captionsetup{justification=justified, singlelinecheck=false, margin=6pt} 
   \caption{Feature importance for the PID considering: Random Forest impurity importance, XGBoost gain importance, SHAP values (computed on Random Forest), and the SOM-derived class-separation importance. Feature importance values are normalized to a maximum of 1 to visually compare feature rankings.}
  \label{fig:feature_importance} 
\end{figure*}

Figure~\ref{fig:feature_importance} compares the feature ranking produced by SOM\_class\_separation against three established baseline importance measures: RF, XGB and SHAP. As shown, all four methods agree that glucose is the most important feature, and broadly agree in ranking BMI, age, and diabetes\_pedigree in the upper-middle tier while ranking blood\_pressure and skin\_thickness lowest. SOM\_class\_separation deviates most from the tree-based methods on age, for which it assigns the second-highest importance of any feature (0.54) versus a comparatively lower ranking from RF, XGB, and SHAP, and on pregnancies, for which it assigns higher relative importance than RF or SHAP. Despite these individual differences in ranking order, the overall agreement across methods derived from entirely different supports SOM class separation as a meaningful importance measure rather than an artifact of the encoding or grid placement procedure

\begin{figure*}
  \centering
  \begin{subfigure}[b]{0.9\linewidth}
    \includegraphics[width=\linewidth]{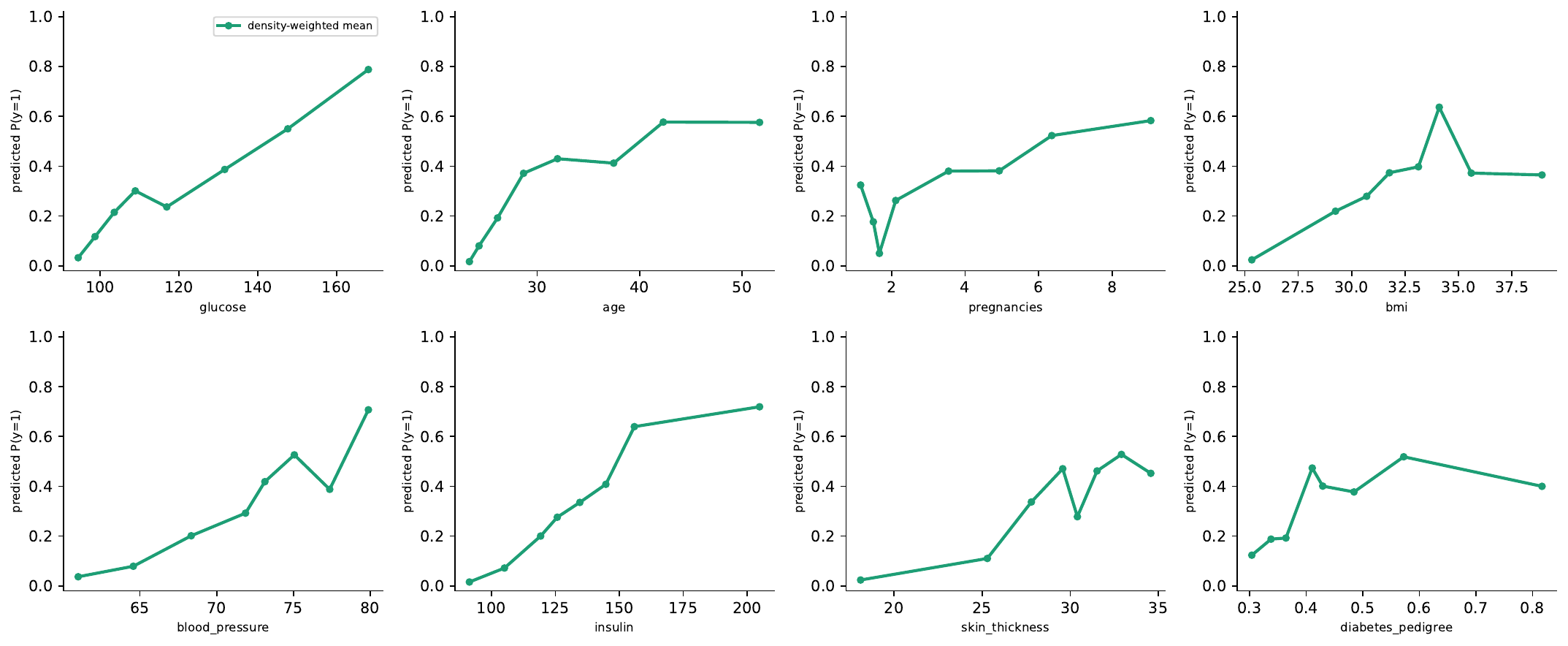} 
    \caption{} 
    \label{fig:gradcam_local:a} 
  \end{subfigure}
  \captionsetup{justification=justified, singlelinecheck=false, margin=6pt} 
   \caption{Prototype-based partial dependence for all features of the PID. Each panel plots the SOM prototype grid's predicted probability against the corresponding feature's component plane value.}
  \label{fig:pdps} 
\end{figure*}

Figure~\ref{fig:pdps} shows the prototype-based partial dependence curve for features of the PID. Four features, glucose, age, blood\_pressure, and insulin, exhibit increasing dependence, with predicted probability spanning roughly the full observed range (0.0 to 0.7–0.8). Glucose in particular shows an almost linear relationship, consistent with its established role as the most direct marker of glycaemic control in this domain. The remaining four features, pregnancies, bmi, skin\_thickness, and diabetes\_pedigree, show non-monotonic curves with a localized dip or spike in an overall increasing trend, and a visibly narrower range of predicted probability than the first group. diabetes\_pedigree shows the flattest and narrowest curve of all features (predicted probability varying only between approximately 0.12 and 0.52 in its observed range), consistent with its role as a comparatively weak indirect risk indicator relative to direct physiological measurements.

\section{Discussion}
\label{sec:discussion}

In this paper, we proposed TabSOM, a tabular-to-image encoding method based on the SOM, and compared it against twelve state-of-the-art tabular-to-image methods across four benchmark datasets. TabSOM achieves competitive classification performance in all evaluated datasets, ranking first or second on every dataset while exhibiting the lowest variance of any method in the comparison. The benchmark comparison reveals two insights. Methods based on deterministic and spatially consistent encodings (BarGraph, Combination, DistanceMatrix, FeatureWrap, and BIE) perform similarly to TabSOM because they produce stable image representations. Methods based on stochastic dimensionality reduction, such as TINTO (t-SNE and PCA variants), DeepInsight (t-SNE and UMAP), IGTD, and Fotomics, perform substantially worse, particularly on smaller datasets such as Pima and QSAR, where the instability of the embedding across folds is reflected in larger standard deviations. TabSOM's component-plane placement avoids this instability by deriving feature positions from the SOM geometry rather than from a dataset-level embedding, producing a layout that is fixed after a single SOM fit.

The interpretability analysis shows that TabSOM provides global explanations through the class-separation importance score and feature interactions via the prototype-based partial dependence. The feature ranking based on class-separation importance identifies features whose component planes are spatially separated by class label, features whose high or low values map systematically to different regions of the SOM grid depending on the outcome. In PID, glucose, age and BMI rank highest according to class-separation importance, consistent with their relevance as relevant factors associated with diabetes. The feature ranking comparison performed with Random Forest, XGBoost, SHAP, and SOM class-separation showed that the SOM-derived ranking agrees moderately with the model-based rankings, with the strongest agreement on the top-ranked features. This suggests that the SOM grid structure captures complementary information, providing interpretability of which features are globally important.

Additionally, the prototype-based partial dependence curves reveal the feature's effect on the classifier's output. For the PID, glucose showed a near-linear positive relationship across its observed range, consistent with its role as a glycaemic marker. Age exhibits a steep rise through the 30s and 40s before plateauing and diabetes\_pedigree produces the flattest and narrowest curve of all eight features, consistent with its status as a comparatively weak risk indicator. Several features (pregnancies, BMI, and skin thickness) showed localized non-monotonicities that the spatial overlay diagnostic traces to low-density regions of the SOM grid or to co-location with higher-ranked features, rather than to independent physiological effects.

In the literature, a single work has explored SOM in tabular-to-image methods. It encodes each sample as a proximity activation map over the SOM prototype grid, measuring Gaussian-weighted distances between the sample and every prototype in RBF kernel space, which produce images where pixel intensities reflect manifold position rather than individual feature values. TabSOM differs in three main dimensions: \textit{(i)} it derives explicit per-feature canvas positions from SOM component planes via Hungarian assignment, producing spatially interpretable images in which each feature occupies a fixed region; \textit{(ii)} it adds a relational edge channel encoding pairwise feature interactions not representable in any marginal-value image; and \textit{(iii)} it provides interpretability tools (class-separation importance, prototype-based partial dependence) validated against established baselines, rather than qualitative activation-pattern inspection. 

Beyond classification performance, interpretability analysis shows that the features identified by TabSOM align with established clinical knowledge, suggesting that TabSOM captures meaningful domain-relevant patterns. This property is especially valuable in high-stakes domains where transparency is as important as predictive performance. Future work will explore integration with multimodal data sources and systematic comparisons with established tabular interpretability methods. Future work will explore comparisons between Grad-CAM and established post-hoc methods for tabular data, such as SHAP, to better understand their strengths and limitations. Finally, future research may investigate the use of alternative architectures, such as vision transformers, to further enhance the representation learning capabilities of TabSOM-generated images.

\section{Conclusions}
\label{sec:conclusions}

This paper proposed TabSOM, a tabular-to-image encoding that uses the SOM to provide both a topology-based feature placement and a relational graph over features, rendered as a multi-channel image that separates marginal feature values from pairwise interactions. We benchmarked against twelve existing tabular-to-image methods on four public datasets. TabSOM ranked first and second on every dataset and exhibited the lowest standard deviation across all methods. TabSOM provides two interpretability tools, including a prototype-inspired partial dependence plot, and the class-separation importance score, to provide feature importance and ranking, and the dependence analysis. A comparative analysis of feature importance with SHAP, and RF showed agreement on the top-ranked features, consistent with the SOM grid capturing structural patterns beyond those encoded by tree-based impurity measures.






\ifCLASSOPTIONcaptionsoff
  \newpage
\fi



%

\bibliographystyle{vancouver}
\bibliography{bare_adv_literature}



%








\end{document}